\documentclass[10pt,twocolumn,letterpaper]{article}

\usepackage[pagenumbers]{cvpr} % To force page numbers, e.g. for an arXiv version

\usepackage{amsmath}
\usepackage{multirow}
\usepackage{caption}

\newcommand{\modelName}{Hoi3DGen} % name of our method
\newcommand{\mat}[1]{\mathbf{#1}}
\newcommand{\set}[1]{\mathcal{#1}}
\newcommand{\vect}[1]{\mathbf{#1}}

\definecolor{ForestGreen}{RGB}{14,109,14}

\definecolor{myPink}{HTML}{E91E63}  % Pink

\definecolor{lightBlue}{HTML}{0277BD}

\definecolor{lightred}{HTML}{FF474D}

\definecolor{cvprblue}{rgb}{0.21,0.49,0.74}
\usepackage[pagebackref,breaklinks,colorlinks,allcolors=cvprblue]{hyperref}

\usepackage{graphicx}
\usepackage[percent]{overpic}

\def\paperID{} % *** Enter the Paper ID here
\def\confName{CVPR}
\def\confYear{2026}

\title{
\modelName{}: Generating High-Quality Human-Object-Interactions in 3D
}
\author{
Agniv Sharma$^{1,4,\dagger}$ \quad
Xianghui Xie$^{1,2,3,\dagger}$\quad
Tom Fischer$^4$ \quad
Eddy Ilg$^4$ \quad 
Gerard Pons-Moll$^{1,2,3}$ \\
{\centering {\small $^1$University of T\"ubingen \quad $^2$T\"ubingen AI Center \quad  $^3$Max Planck Institute for Informatics \quad $^4$Technische Universit\"at N\"urnberg} } \\
{\small\href{https://virtualhumans.mpi-inf.mpg.de/hoi3dgen/}{https://virtualhumans.mpi-inf.mpg.de/hoi3dgen/}}
}

\begin{document}
\twocolumn[{%
\renewcommand\twocolumn[1][]{#1}%
\maketitle
\begin{center}
    \centering
    \captionsetup{type=figure}
    \includegraphics[width=0.96\textwidth]{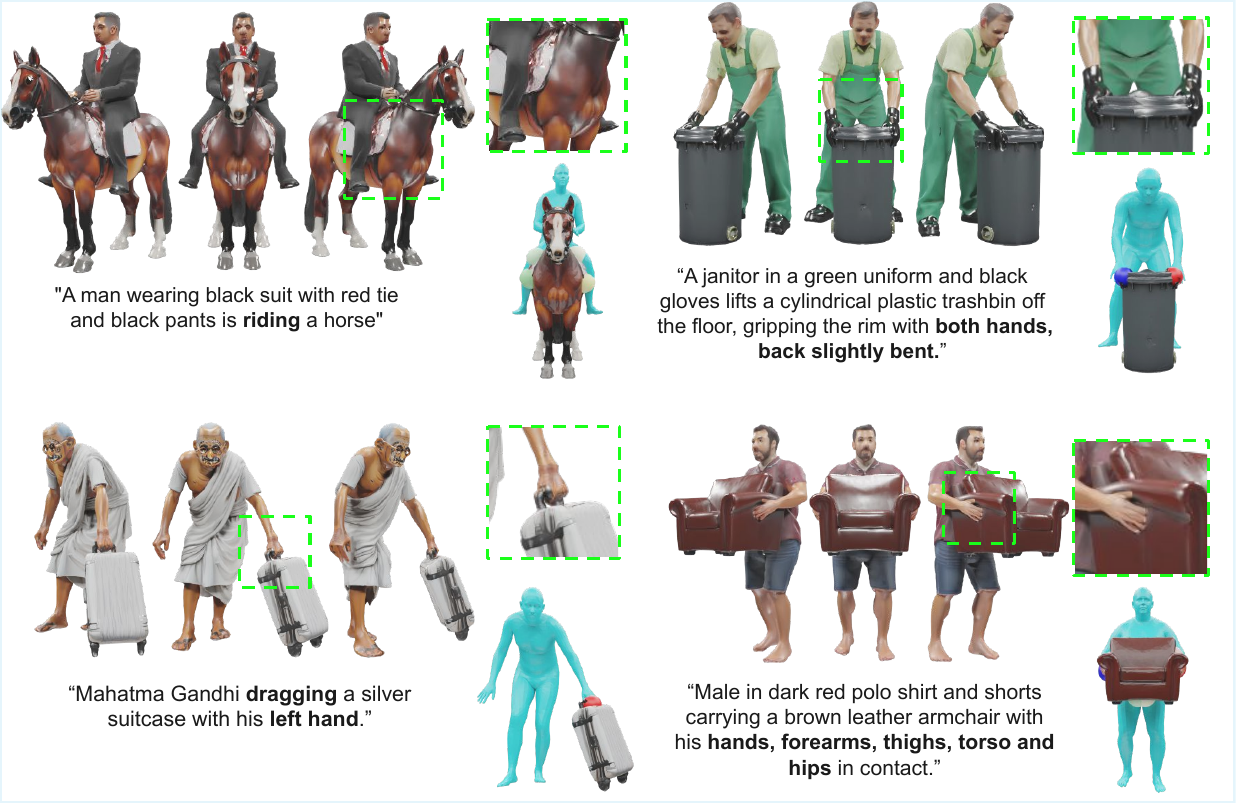}
    \captionof{figure}{
    Given detailed text descriptions of human, object and their interactions, \modelName{} generates high quality textured human and object meshes that follow precisely the contact semantics, together with an aligned animatable SMPL model. 
    % We present the first text-to-3D generation method that generates textured 3D models with high-quality human-object interaction from text prompts. The method outputs segmented meshes for humans and objects as well as an animatable SMPL model.    
    %Given a detailed interaction description, \modelName{} generates 3D  meshes precisely following the specified body part contacts. This is achieved by distilling the strong interaction priors in existing foundation models with our automatically 
    %generated dataset containing pairs of text and 3D interaction\EI{how is the 3D interaction represented? meshes?}. Note how well our model generalizes to diverse humans, object categories and interaction types, while the used\EI{which used dataset? the one that we generate? That seems like a contradiction} interaction dataset has limited variation.
    % \EI{The figure appears a bit crowded, leaving slighty more space between the subfigures would be good.}\EI{|The figure may be too big, leaving away one row would also work.}
    % \TF{The zoom-in is essentially the same size as the original bracket. Consider zooming in more to make it more detailed.} \XH{I feel there is too much white space between figure and caption, and the white space between two image rows can be squeezed a bit more. move the text input above the figure and write `Input: "$<\text{the text prmpts}>$", between text and figure, add text `High quality 3D human-object interaction'}
    % \AG{I think above two comments are contradictory so we should stick with what we have as it seems to be a nice balance, I can change font size}
    }
    \label{fig:teaser}
\end{center}%
\vspace*{4mm}
}]

% --- ADD THIS BLOCK TO RESTORE FOOTNOTES ---
{
  \renewcommand{\thefootnote}{\relax}
  \footnotetext{$^\dagger$Equal contribution.}
  % \footnotetext{$^*$Corresponding author} % Uncomment if you need a footnote for the star too!
}
% -------------------------------------------

\begin{abstract}
Modeling and generating 3D human–object interactions from text is crucial for applications in AR, XR, and gaming. Existing approaches often rely on score distillation from text-to-image models, but their results suffer from the Janus problem and do not follow text prompts faithfully due to the scarcity of high-quality interaction data. 
We introduce \modelName{}, a framework that generates high-quality textured meshes of human-object interaction that follow the input interaction descriptions precisely. We first curate realistic and high-quality interaction data leveraging multimodal large language models, and then create a full text-to-3D pipeline, which achieves orders-of-magnitude improvements in interaction fidelity. Our method surpasses baselines by 4–15$\times$ in text consistency and 3–7$\times$ in 3D model quality, exhibiting strong generalization to diverse categories and interaction types, while maintaining high-quality 3D generation.
\end{abstract}    
\section{Introduction}
\label{sec:intro}
% \TF{General comment for this section: I think there is a lot of repetition of abbreviations (3D and HOI in particular). I know this is hard to avoid given the nature of the paper, removing/rewriting would improve the flow a lot imo. Also, currently foundation and foundational models are used interchangeably. I would stick to "foundation" model only.}
Modeling human-object-interaction is highly important for gaming, virtual and augmented reality.
% gaming, virtual reality and robotic applications. 
Generating such interactions as 3D models from text prompts is particularly interesting, since the manual creation of interacting humans and objects is laborious. 
%
%as it is very laborious to manually create 3D assets with both detailed human, object and accurate interaction configurations. 
Despite its significance, this problem has been overlooked, and most methods focus on either human-only~\cite{hong2022avatarclip, cao2023dreamavatar} or object-only~\cite{xiang2024trellis, instant3d} generation. Only a few methods~\cite{dai2024interfusion, chen2024comboverse, zhu2024dreamhoi, zhang2025interactanything} have pioneered this field with score distillation sampling (SDS). While showing promising results, SDS generations are unreliable at inference and often lead to unnatural poses and low-quality 3D interactions with the Janus problem.

A major challenge for text-to-3D interaction generation is the lack of paired text captions and 3D interaction data. Existing 3D interaction datasets~\cite{xie2023template_free, bhatnagar22behave, GRAB:2020, li2023omomo} feature interactions with natural poses and high-quality contacts, but typically cover only a limited set of object categories or lack fine-grained textual descriptions. Consequently, current approaches often adopt training-free SDS-based pipelines that leverage powerful image diffusion models~\cite{dai2024interfusion,zhu2024dreamhoi, zhang2025interactanything}. While the underlying models were trained on billions of images, these models do not generalize well for interactions and produce implausible results. 
%
%their limited 3D understanding ability often leads to physically implausible interactions, bad contacts, or artifacts like the Janus problem.

In this paper, we introduce \textbf{\modelName{}}, a framework that generates high-quality 3D-human-object-interaction, while also maintaining strong generalization ability. Our key idea is to automatically create high-quality text descriptions for 3D interactions and fine-tune a model to generate the compositional interaction without losing its own capability for generating diverse humans and objects. Specifically, we propose an automatic interaction labelling pipeline that leverages multimodal large language models to generate high-quality and detailed descriptions for 3D human object interaction. We decompose the complex interaction captioning task into simpler subtasks that describe appearance, action, and body parts in contact, respectively, and fuse the results together to generate the final caption. Based on automatically generated data, we design a text-to-3D generation method that equips the text-to-image model with view conditioning, lifts the 2D image to a 3D mesh, and segments and registers the 3D mesh with the SMPL model to obtain semantic contacts. 
% Experiments show that \modelName{} surpasses baselines by 4–15$\times$ in text to 3D consistency and 3–7$\times$ in 3D model quality. Ablation studies further highlight the importance of our design choices. 
In summary, our main contributions are:
\begin{itemize}
    \item We present an automatic data annotation pipeline that generates high-quality detailed text captions for 3D-human-object-interactions by decomposing the complex caption generation into subtasks, solvable by open-source multimodal large language models.     
    \item We introduce a text-to-3D pipeline that generates high-quality interacting segmented human-object meshes with an aligned animatable SMPL model. 
    % With the dataset, we establish a text-to-3D pipeline for generating high-quality human-object-interactions that outputs segmented meshes for human and object, as well as an animatable SMPL model.
    % \pagebreak 
    \item We demonstrate that \modelName{} surpasses baselines by 4–15$\times$ in text to 3D consistency and 3–7$\times$ in 3D model quality.
    % We show that the results of existing models are implausible and of low quality, and for the first time present a high-quality text-to-3D generation pipeline. 
\end{itemize}
Our code, data, and pretrained models will be released. 

\section{Related work}
% \EI{In general: it is not correct grammar to start a sentence with [] (e.g. "[52] proposes X", instead you need to mention the author names in this case. E.g. "Someone et al.[52] proposes X"}\EI{using caps in the tiles below is not consistent}

\noindent\textbf{Text-to-3D generation.} For general objects, recent methods can be classified as Score Distillation Sampling (SDS)-based or learning-based. SDS-based approaches like DreamFusion \cite{poole2023dreamfusion}, ProlificDreamer \cite{wang2023prolificdreamer} and more \cite{melaskyriazi2023realfusion, chen2023fantasia3d, tang2023make, lin2023magic3d} distil 3D objects from pre-trained 2D image diffusion models~\cite{rombach2021highresolution}. 
% SDS-based methods are generalizable\EI{scalable?} as they do not require 3D data for training. However, to make them work effectively, researchers often design task-specific architectures to address the challenges that arise\EI{citations to examples would be good}. In contrast, we show that using unambiguous data samples to guide 2D priors is more effective than architectural modifications.\EI{the last sentence is not clear to me} \textcolor{red}{This is an important point but also we only show it for one specific case, how do I change statement to reflect that and not make it seem like overclaiming?} 
Learning-based approaches either fine-tune image diffusion models to generate novel views conditioned on camera poses \cite{shi2024mvdream, kant2024spad, Xue2024gen3diffusion, xue2024human3diffusion}, or adopt native 3D representations like Triplane \cite{Chan2022ED3D, openlrm}, SDF \cite{hunyuan3d22025tencent}, occupancy \cite{zhang2024clay}, or hybrid ones \cite{xiang2024trellis} to directly obtain 3D. Trained on large-scale datasets \cite{deitke2023objaverse, objaverseXL}, these methods can generate high-quality 3D from text. Orthogonal to these, human avatar creation relies more heavily on SDS due to the lack of high-quality and diverse human data \cite{hong2022avatarclip, cao2023dreamavatar,kolotouros2023dreamhuman,zhang2023avatarverse, wang2024disentangledavatar, liu2024humangaussian}. Recent works \cite{Zhuang2025idol, xue2025infinihuman} propose to distill 2D generation models to create large-scale human data. While showing promising results, these methods consider only humans or only objects and cannot model the complex relationship during interaction. 

% Text to 3D generation, most of them focus on object or human only, some work on scenes. only few consider interaction or compositional objects. 

% but they are slow, and do not have high quality, cannot understand complex interactions. 

\noindent\textbf{Human-object interaction.}
Most existing methods focus on the accurate capture of hand-object~\cite{hasson19_obman, yang2021cpf}, full-body-object~\cite{xie22chore, xie2024rhobin} or human-scene~\cite{PROX:2019, Yi_MOVER_2022, yalandur2025physic} interactions from images~\cite{xie2023template_free} or videos~\cite{fan2024hold, xie2023vistracker, xie2024InterTrack, xie2026cari4d, chen2025human3r}. Based on the captured interaction motion data \cite{bhatnagar22behave, li2023omomo, GRAB:2020}, several works learn to generate HOI motion sequences conditioned on signals like human or object position~\cite{li2023omomo, wu2024thortexthumanobjectinteraction}, past motion sequence~\cite{xu2023interdiff}, and text \cite{diller2023cghoi, li2023controllable, peng2023hoi, xu2024interdreamer, wu2025humanobjectinteractionhumanlevelinstructions}. Despite impressive performance, these approaches are typically limited to non-clothed SMPL body meshes and object templates from the training datasets, restricting generalization. 
Another line of work synthesizes interactions from provided human and object meshes and optimizes human poses via SDS~\cite{zhu2024dreamhoi, zhang2025interactanything}, with \cite{zhang2025interactanything} further improving contacts using open-set affordances and LLMs. 
However, they require user-supplied meshes and cannot generate fully-textured HOIs from text. 
The most related work \cite{dai2024interfusion}, employs SDS to optimize separate human and object NeRFs and fuses them into a combined scene, but suffers from the Janus problem, noisy textures, and severe interpenetrations, making segmentation difficult and lacking contact control. In contrast, our method produces realistic human–object interactions with high-quality textures, accurate contacts, and also enables accurate human-object segmentation.

\noindent\textbf{3D datasets with text annotation.} With the great progress in large language models (LLM), automated text annotation for various 2D \cite{schuhmann2022laion5b, xue2025infinihuman} and 3D data \cite{zhou2018realestate10k, li2026frankenmotion} has become a viable option. 
For objects, CAP3D~\cite{luo2023CAP3D} proposed a scalable pipeline using vision language models and annotated the Objaverse dataset~\cite{objaverseXL}. Follow-up work \cite{luo2024view_sel} further improves the accuracy by robustly selecting better views for annotation. These works output only one description of the object, while Marvel-40M+ \cite{sinha2024marvel} introduces multi-level annotations, which allow control at different complexity levels. Orthogonal to these, PoseScript \cite{delmas2022posescript} develops a procedural pipeline to assign text labels based on axis angles and further train a model for automatic human-pose-to-text annotation. However, these methods handle objects and humans separately. Some works ~\cite{peng2023hoi, xu2025interact, wu2024thortexthumanobjectinteraction} annotated interaction motion with text manually, which is not scalable. Another work \cite{yang2024fhoi} utilizes GPT-4V for detailed HOI labeling, but its dependence on proprietary models introduces cost constraints, and its lengthy descriptions deviate from natural human phrasing. In contrast, we propose a fully automatic and scalable pipeline based on open-source models to annotate complex interactions through natural language.

% Datasets with text annotations: Cap3D, Marval1M+, but none of them deal with interactions. 

% text BEHAVE: https://arxiv.org/pdf/2403.11208, manual annotation which is not scalable. 

\section{Method}
\begin{figure*}[th]
    \centering
    \includegraphics[width=1.0\linewidth]{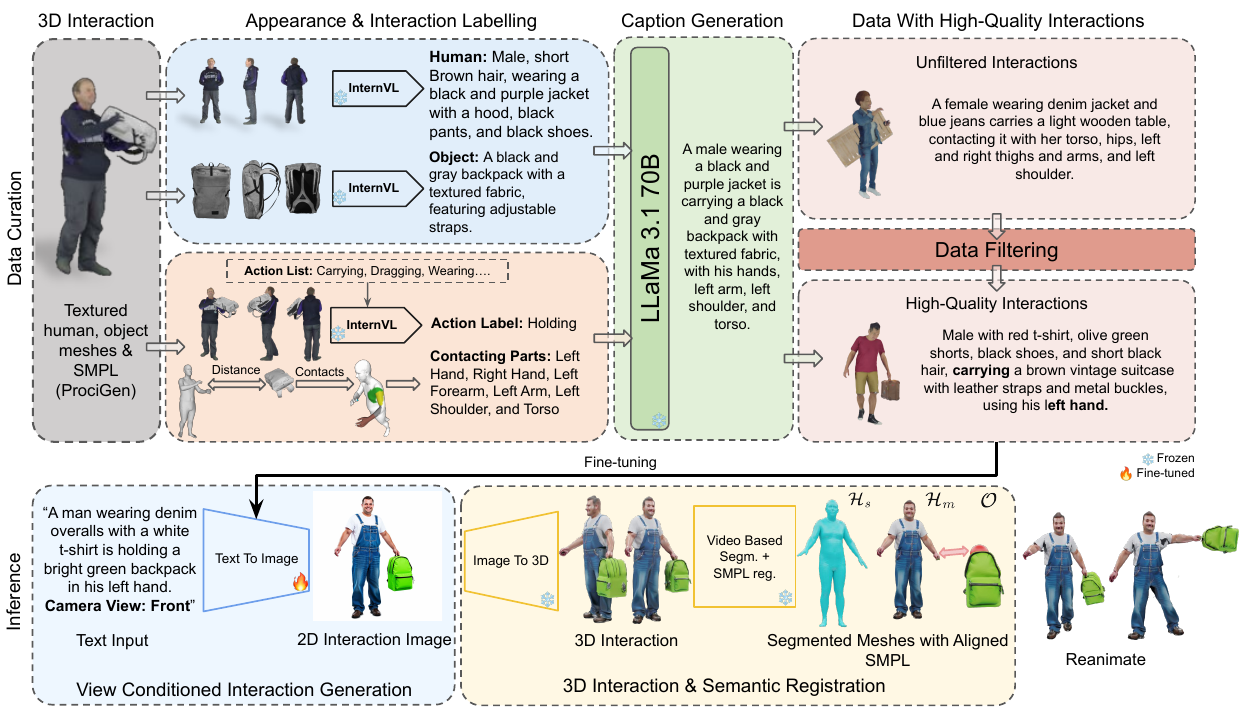}
    \caption{\textbf{HOI3D framework overview.} 
    \textbf{Top:} 
    We first leverage the existing multimodal foundation model InternVL~\cite{chen2024internvl2} to perform decomposed annotation of human, object, and human-object-interaction of samples from the ProciGen~\cite{xie2023template_free} dataset. We then use LLaMa~\cite{grattafiori2024llama3} to create a final detailed caption for the sample. 
    \textbf{Bottom:} We leverage our data consisting of high-quality and diverse human-object-interactions to fine-tune an existing text-to-image model. Subsequently, we establish a pipeline to reconstruct high-fidelity textured 3D meshes. The output of our final text-to-3D inference pipeline consists of segmented meshes for the human and object, as well as an animatable SMPL model. 
    }
    \label{fig:DistiGen}
\end{figure*}

% An overview of our method is provided in~\cref{fig:DistiGen}. We will first describe the critical data curation and subsequently our text-to-3D inference pipeline.  

We present \modelName{}, a framework to obtain high-quality 3D models of humans interacting with objects. An overview of our method is provided in~\cref{fig:DistiGen}. 

First, we introduce an automatic data annotation pipeline that produces detailed text captions for 3D-human-object-interaction (\cref{subsec:data-curation}). We then adopt a text-to-image and 2D to 3D lifting paradigm for 3D generation. We propose a view-conditioned image generator that synthesizes interaction images from text (\cref{subsec:text2image}) and lifts them into high-quality textured 3D meshes, which are then semantically segmented into human and object components, and aligned with the SMPL model for consistent semantics and animation (\cref{subsec:2d-to-3d}).

\subsection{Data Curation}\label{subsec:data-curation}

Given a 3D human–object-interaction mesh, we aim to automatically generate captions that describe appearance, actions, and contacts. For scalable automation, we decompose the task into subtasks that multimodal LLMs can solve efficiently and reliably. As is common in many interaction datasets~\cite{bhatnagar22behave, zhang2023neuraldome, xie2023template_free} and also the case in ProciGen, we assume that the 3D interaction is represented as a textured SMPL~\cite{loper2015smpl} model for the human and a separate textured mesh for the object, see \cref{fig:DistiGen} top left. 
We then divide the interaction annotation into 1.) appearance labeling, 2.) interaction labeling and 3). caption generation.
%(in our experiments we chose InternVL-2.5~\cite{chen2024internvl2})

\noindent\textbf{Appearance Labeling.} To label humans and object, our method begins by rendering them separately into four orthogonal views: front, back, and either side. 
We then use these views as queries for the multimodal LLM InternVL~\cite{chen2024internvl2} and prompt it to describe the attributes shown in the image. 
For human labeling, we focus on clothing, hairstyle, and footwear. For object labeling, we annotate attributes such as color, texture, and overall structural form.

\noindent\textbf{Interaction Labeling.} 
A detailed interaction description should define the \textit{what} and \textit{how} of the interaction.
We query the InternVL with the four views of human-object interaction renderings and ask it to describe the type of interaction by picking a label from a predefined list of possible actions of the shown object category.
Alternatively, one could let InternVL propose actions, but we found that this leads to poor annotation quality due to hallucinations or ambiguous wording for semantically similar actions.
Addressing the \textit{how} is also crucial, since the same action can produce very different interactions depending on the contact points (i.e. \textit{holding} a basket in the \textit{right} or \textit{left} hand). 
To integrate this information, we analyse contact points of the SMPL mesh with the object and filter out body parts whose distances are smaller than 4cm as the contacting parts.

\noindent\textbf{Combined Caption Generation.} To generate the final caption, we integrate human, object, action, and interaction labels, along with the object category, using LLaMA 3.1 (70B) \cite{grattafiori2024llama3}. The strong generative capability of the model helps us in obtaining high-quality, natural, and detailed text captions.

\noindent\textbf{Filtering.}
We apply our data annotation method to the entire ProciGen dataset~\cite{xie2023template_free}, yielding over 750k pairs of 3D models and captions.
Although ProciGen features large object shape diversity, the human appearances and interaction types are limited (100 subjects and 18 categories, respectively).
More importantly, many ProciGen interactions involve multiple simultaneous contacts, which leads to a less rich training signal when finetuning, since the model has to distinguish contacts such as left and right part labels.
To avoid forgetting during the finetuning, we aim to minimize the training iterations and therefore curate a small subset of high-quality and diverse samples.  

We separate interactions according to $K=8$ frequent but distinct contact configurations: on back, right hand, left hand, right leg, left leg, both hands, no contact, and others.
% \EI{where does 8 come from?}, 
The goal is to construct disjoint subsets $\set{D}^k=\{x^k_i\}$, where each sample $x^k_i=(t, \set{I})$ consists of text description $t$ and renderings $\set{I}$ that \emph{only} have contact configuration $k$.
For instance, \textit{right hand} includes only those with contact limited to the right hand and forearm. 
Then, we filter each subset by removing data where the object is either significantly overlapping with the human or is far away from both the human and the ground. 
%, we further remove examples where multiple contact parts are involved in the interaction. 
We then group the remaining data based on action-contact pairs and remove samples where the action clearly mismatches the contact (i.e. a sample with contact point \textit{right hand} and action \textit{kicking}).
Finally, we arbitrarily select $50$ samples from each subset $\set{D}^k$, leading to 400 remaining high-quality 3D HOI instances.
% \TF{The 8 contact points are arbitrarily chosen I believe?}

\subsection{View-Conditioned 2D Interaction Generation}\label{subsec:text2image}

For text-to-image generation, we build on top of the latent diffusion model SANA~\cite{xie2024sana}, which can already generate high quality human-only or object-only images but cannot follow precisely the interaction prompts.
%
%follow a latent diffusion paradigm due to its ability to precisely follow the text description when it comes to human appearance, object style, and scene types. 
%To also accurately reflect the intricate spatial relationships and contacts during human-object interaction we finetune a pre-trained SANA~\cite{xie2024sana} model with our curated dataset. 
% As such text-to-image generation models can already generate rich distributions of humans and objects alone, 
A key discovery of our work is that fine-tuning on the $400$ high-quality and diverse interaction samples is sufficient to adjust the learned representation, enabling the model  to generate high-quality human-object interactions while maintaining its original power of generating diverse humans and objects. 

Another key innovation to enable lifting to high-quality 3D models is to introduce a conditioning of the text-to-image model on the camera view angle.
%This ensures that the interaction concept from the text is clearly visible, greatly simplifying the workload as occlusions are minimized. 
Specifically, we append a view description $t_v$ to the interaction prompt $t$  to generate the 2D interaction image $\mat{I}$ using our latent diffusion model $\epsilon_\theta$.
We allow control of three distinct view angles $t_v\in \{\text{front, left diagonal, right diagonal}\}$ that avoid occlusion for most interaction types. These correspond to rendering 3D interaction meshes from azimuth angles of $0^o, -45^o, +45^o$ respectively. 
We then finetune the model using the standard diffusion loss~\cite{ho2020ddpm}: 
\begin{equation}
\begin{aligned}
&\mathcal{L} =
\mathbb{E}_{x_0,\, \epsilon,\, \sigma}
\left[
\left\|
\epsilon - \epsilon_\theta(z_t, \sigma, \text{cond})
\right\|_2^2
\right], 
\end{aligned}
\label{eq:sana-loss}
\end{equation}
where $\epsilon \sim \mathcal{N}(0, I),\,
\sigma \sim \mathcal{U}(0, 1)$,  and
$ \text{cond}=(t, t_v)$. 
Finally, since training on the highly correlated finetuning data will inadvertently bias the model and reduce output variety, we improve overall texture quality and fidelity in the generated output by retexturing with the Flux model \cite{flux2024}.  
% \AG{This is a technical detail but the main reason the model decreases in quality is because procigen renderings are low quality as they are just textured meshes, but we can also write this I guess}

We show in \cref{tab:view-cond-sample} that our view-conditioned sampling produces more accurate 3D contacts and our fine-tuning significantly improves the existing text-to-image model for interaction image generation in \cref{tab:text-to-2d-eval}. 

\subsection{3D Interaction and Semantic Registration}\label{subsec:2d-to-3d}

\noindent\textbf{3D Interaction Generation.}
Given the high-quality image of our finetuned diffusion model, we can obtain a 3D mesh using a large-scale image-to-3D model.
In our experiments, we use Hunyuan3D~\cite{hunyuan3d22025tencent} due to the observation that it can generate diverse shapes and interactions if the 2D image is of high quality. Our view control allows us to sample three images of the same interaction type, where at least one view has the full interaction visible from which Hunyuan3D can easily generate high quality and complete 3D mesh. Hence, we directly apply Hunyuan3D to all generated images and obtain three different textured 3D interactions. The user can either pick the best one manually or we perform automatic selection based on segmentation and contact semantics, which we describe next. 
% output the textured 3D interaction mesh obtained by running the image-to-3D model on our sampled interaction images. 
% Our view control allows us to sample multiple views of the same scene, which is a strong enough signal to guide image-to-3D models to generate high-quality 3D interaction meshes with correct contacts, as shown in \cref{subsec:ablation}. 

% We show in \cref{subsec:ablation} that Hunyuan3D faithfully preserves the 2D contact semantics with our generated images.  

\noindent\textbf{Interaction Segmentation.}
Despite high quality, the 3D generation from Hunyuan3D is a single combined mesh in a normalized space. For practical applications, one also needs to understand the semantics, such as which part the human is and where contacts happen. 
To this end, we propose to segment the mesh and register a SMPL~\cite{loper2015smpl} body model that provides semantics and allows further animation. 

Given the combined 3D mesh $\set{M}$ produced by Hunyuan3D, our objective is to separate it into two semantically meaningful components: the human mesh $\set{H}_m$ and the object mesh $\set{O}$.
To this end, we render $\set{M}$ into a video sequence along a smooth camera trajectory that spans elevations in $[-60^\circ, 60^\circ]$ and a full $360^\circ$ azimuth.
We then apply the open-vocabulary video segmentation model Grounded-Segment Anything 2 (GSAM2)~\cite{kirillov2023segany, ren2024groundedSAM, liu2023groundingDINO} to obtain temporally consistent binary mask sequences $\{\mat{M}_i^h, \mat{M}_i^o\}_{i=1}^N$ of human and object, for each rendered view $i$. We input \textit{person} and the target object category to prompt GSAM2 for human and object segmentation respectively. 
% Since the human mesh is reconstructed from SMPL, we prioritize accurate object segmentation. 

We aggregate the 2D masks to segment 3D mesh vertices based on vertex visibility and majority voting. 
% Given object mask $\mathbf{M}_i^o$ and depth map $\mathbf{D}_i$ rendered in view $i$, we first compute the visibility of vertex $\vect{v}$ by computing 
For each rendered view $i$, we have access to the object mask $\mathbf{M}_i^o$ of GSAM2 and the depth map $\mathbf{D}_i$ and camera parameters of the rendering process.
A vertex $\vect{v}$ is considered \textit{visible} in view $i$ if it projects inside the image and passes a z-buffer consistency check:
\begin{equation}
    \text{vis}_i(\vect{v})=\mathbf{1}[-\delta \leq z_i(\vect{v}) - \mathbf{D}_i(\pi_i(\vect{v})) \leq \delta],
\end{equation}
% \XH{is this absolute?}
where $\delta$ is a depth tolerance, $\pi_i(.)$ is the projection function of frame $i$, and $z_i(\vect{v})$ denotes the projected depth of $\vect{v}$ to view $i$.
For the set of views where $\vect{v}$ is visible, we compute the fraction in which the vertex lies inside the object mask.
Let $\mathcal{V}(\vect{v})$ denote the set of visible views of $\vect{v}$.
A vertex is then assigned the binary object label if this fraction exceeds a certain threshold $\tau$ (empirically, 0.5):
\begin{equation}
l(\vect{v})=
\begin{cases}
    1, & \text{if } \frac{1}{|\mathcal{V}(\vect{v})|}\sum_i \mathbf{M}_i^o[\pi_i(\vect{v})] > \tau, \\
    0, & \text{otherwise,}
    \label{eq:object-segm}
\end{cases}
\end{equation}
Finally, the mesh $\mathcal{M}$ is split into the human and object components $(\mathcal{H}_m, \mathcal{O})$ according to these per-vertex labels.

\noindent\textbf{SMPL Registration.} To obtain semantics for the generated interaction, we register a SMPL~\cite{loper2015smpl} body model $\set{H}_s$ to our segmented human mesh $\set{H}_m$. 
Since the human mesh is often incomplete, off-the-shelf human registration methods do not work well for our setup, since they require complete scans \cite{xie2023template_free} or were trained on limited poses~\cite{li2025etch}.
To solve this issue, we introduce a simple yet effective approach to obtain an aligned SMPL mesh. 
First, we apply CameraHMR~\cite{patel2024camerahmr} to a arbitrary rendering with zero elevation to identify the front view of the mesh.
Then, we render the mesh from the front to get a higher quality SMPL by reapplying CameraHMR.
The SMPL model is then scaled and translated to match the center and scale of our mesh output.
Finally, scale, global translation, and rotation (7DoF) are refined with a few rounds of Chamfer distance-based optimization. 
%We show in \cref{subsec:ablation} that our approach works more robustly than the state-of-the-art method ETCH~\cite{li2025etch}. 

%With our accurately aligned SMPL model, one can further animate the generated interaction or reason where the contacts happen, which is useful to  evaluate the faithfulness of the generated interaction. 

\begin{figure}[t]
    \centering
    \includegraphics[width=1.0\linewidth]{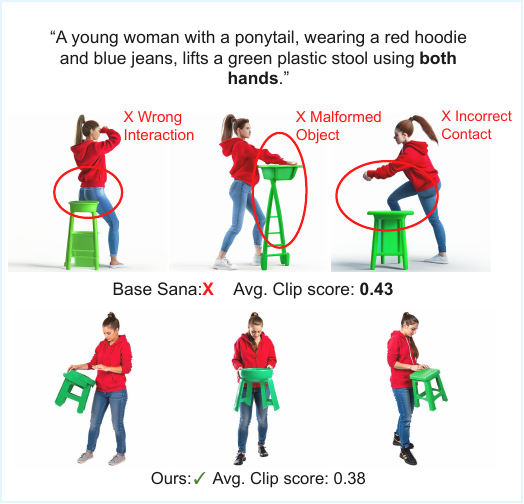}
    \caption{
    \textbf{Analysis of the CLIP score.} While our model clearly generates images that follow input interaction descriptions more precisely than SANA~\cite{xie2024sana}, the CLIP score indicates the opposite, rendering it unusable as a metric for our task.  
    % We show that CLIP's utility is limited for judging human-object interaction images. CLIP clearly misunderstands interaction, contacts, and quality. 
    % \XH{the red texts are too small while the titles in the bottom is too big. and the prompt fornt and style is not consistent with other figures.}
    }
    \label{fig:clip_fail}
\end{figure}

\section{Experiment}
\label{sec:experiments}
% \AG{Please arrange the tables I am bad at this}
\subsection{Experimental Setup}
\label{subsec:experimental_setup}
\textbf{Implementation Details.}
For our text-to-image pipeline, we fine-tune the pretrained \textit{Sana} model at a resolution of 1024×1024. The fine-tuning process is carried out on four H100 GPUs over 24 hours, with each GPU handling a batch size of 4, yielding an effective batch size of 16.
All inferences are performed on a single A100 GPU. 
% \XH{this can go into supp.}

% For image-to-shape generation, we use \textit{Hunyuan3D-DiT-v2.0} for geometry and \textit{Hunyuan3D-Paint-v2.1} for texture synthesis. The \textit{2.0} model provides more stable geometric results in our experiments, while \textit{2.1} yields higher-quality, PBR-compatible textures. 
% \TF{For image-to-3D generation we use Hunyuan3D~cite[github or paper, whatever exists].}

\noindent\textbf{Evaluation Metrics.} We evaluate model performance in two aspects: consistency to input text prompt and quality of the generated 3D interactions. For text consistency, we report GPT score, CLIP score, and contact accuracy. The \textbf{GPT score} is defined in the same way as in InterFusion~\cite{dai2024interfusion}, where GPT-4V is prompted to select the one that is most consistent with the input text among the generations from different methods. Each 3D generation is rendered into 4 orthogonal views as the input to GPT-4V. \textbf{CLIP score} averages CLIP similarity \cite{radford2021clip} across eight rendered views. While we report this standard metric for completeness, we find that CLIP is poorly suited for fine-grained interaction recognition as we show in \Cref{fig:clip_fail}. We also report \textbf{contact accuracy} to evaluate if our 3D interactions faithfully follow the contacts defined in input prompts. Using our SMPL registration to the 3D interaction, we can segment out the body parts that are mentioned in the prompt and consider it in contact if its minimum distance to the object is smaller than 4cm, which is the same as \Cref{subsec:data-curation}. We then calculate the percentage of contacts that correctly follow the contacts defined in prompts as the contact accuracy. 

For 3D quality, we also instruct GPT-4V to select the best one among generations from different methods. In this case, however, we only prompt the model to consider the visual quality alone without providing the prompt to generate the 3D. We additionally conduct user studies to evaluate the text consistency and 3D fidelity, detailed later. 

\noindent\textbf{Interaction Prompts.} We ask ChatGPT to generate 100 prompts describing humans and objects in various interaction scenarios and use these prompts as input to produce our 3D interactions. These prompts cover the most general interactions and are used to report the GPT score in text consistency, quality, CLIP score, and the user study. For contact accuracy, we use ChatGPT to generate 60 prompts focusing specifically on the most important body parts for interaction: hands and feet.

\subsection{Text-to-3D Interaction Generation}
\label{subsec:3d_baseline}
Our \modelName{} allows generation of 3D human-object interaction (HOI) with contacts controlled precisely by text descriptions. We compare our method against TRELLIS~\cite{xiang2024trellis}, a state of the art method for text to general 3D object synthesis, and InterFusion~\cite{dai2024interfusion}, current state-of-the-art method for text-to-3D textured HOI generation. 

\maketitle
\begin{figure*}[t]
    \centering
    \includegraphics[width=\textwidth]{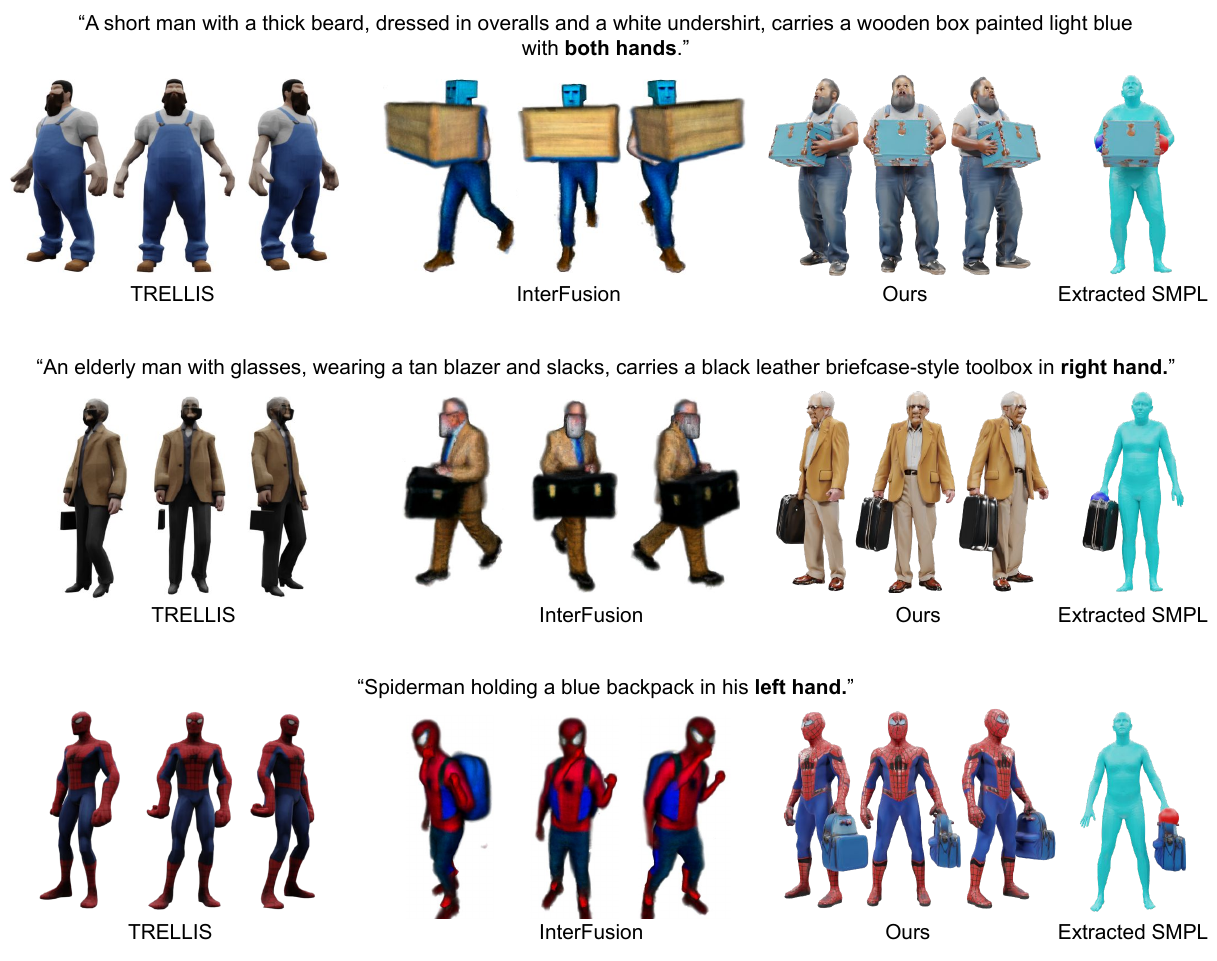}
    \caption{\textbf{Qualitative comparison for text to 3D generation}. InterFusion \cite{dai2024interfusion} is based on Score Distillation Sampling and hence is slow and produces low-quality 3D due to the well-known Janus problem. TRELLIS \cite{xiang2024trellis} is a learning based native 3D generation method, hence it can produce better 3D but is not interaction-aware. Our method faithfully follows the text prompts, especially the detailed body contact specifications. Our contacts are highlighted with spheres coloured based on contacting body parts. 
    }
    \label{fig:qualitative_compare}
\end{figure*}
% For the 3D baseline evaluation we evaluate our model against two baselines.  The first baseline, InterFusion \cite{dai2024interfusion}, represents the current state-of-the-art in fully textured Human-Object Interaction (HOI) generation. The second baseline, TRELLIS \cite{xiang2024trellis}, is the leading method for text to general 3D object synthesis.

\paragraph{Quantitative evaluation.} We report the text consistency and 3D quality metrics in \Cref{tab:text-to-3d-eval}. It can be seen that our model significantly outperforms baselines in GPT scores. TRELLIS generates 3D interaction as a full mesh, hence it cannot reason about the semantic contacts between human and object. InterFusion generates the interaction conditioned on a SMPL mesh, yet the interaction is not aligned with the original human mesh, hence it also cannot reason about the contacts. Our model generates 3D interaction together with a registered human body model, and the contact follows precisely the input prompts with an accuracy of 90\%. 
% \subsubsection{Quantitative Evaluation}
% \label{subsubsec:3d_quantitative}

% We also conduct user studies to evaluate text consistency and 3D quality. We randomly select 40 examples where participants are asked to evaluate the 3D quality for the first 20 examples and text consistency for the later 20 examples. For each question, we render 360$^\circ$ view videos of TRELLIS, InterFusion and our generations side by side with random orders. We release the user study to 33 participants and in total our method is preferred by 91.09\% in text consistency and 85.56\% in 3D quality, thereby clearly outperforming all other methods by almost an order of magnitude. Please refer to the Supplemental material for example user study questions. 
As shown in~\Cref{tab:text-to-3d-eval}, to obtain most faithful assessments, we also conduct a user study to assess text consistency and 3D quality on 40 randomly selected examples, 20 for each criterion. Participants view 360° videos of TRELLIS, InterFusion, and our results in random order. Among 33 participants, our method is preferred by 91.09\% for text consistency and 85.56\% for 3D quality, significantly outperforming other methods. We provide the details on the user study in the Supplemental material.

Notably, InterFusion achieves better text consistency than TRELLIS, yet its 3D quality is worse than direct 3D generation from TRELLIS. This is attributed to its reliance on score distillation sampling which suffers from low-resolution results and the Janus problem. Our method generates 3D HOI directly, while also maintaining high quality.

\begin{table}[t]
\centering
\footnotesize
\setlength{\tabcolsep}{3.5pt}
\begin{tabular}{l|cccc|cc}
\toprule
\multirow{2}{*}{Method} &  \multicolumn{4}{c|}{Text Consistency} & \multicolumn{2}{c}{3D Quality} \\ 
& GPT~$\uparrow$ & CLIP~$\uparrow$ & Contact~$\uparrow$ & User~$\uparrow$  & GPT~$\uparrow$ & User~$\uparrow$\\
\midrule
TRELLIS & 0.04 & 0.32  & N/A & 3.44\%& 0.21 & 10.16\% \\
InterFusion & 0.15 & 0.35  & N/A & 5.47\%& 0.00 & 3.28\%\\
\textbf{Ours} & \textbf{0.81} & \textbf{0.42} & \textbf{90\%}& \textbf{91.09\%} & \textbf{0.79} & \textbf{85.56\%}  \\
\bottomrule
\end{tabular}
\caption{\textbf{Quantitative comparison with text-to-3D models.} We compare our method against TRELLIS~\cite{xiang2024trellis}, a general text-to-3D object generation model, and InterFusion~\cite{dai2024interfusion}, a text-to-3D interaction model. Our method outperforms all prior arts in both consistency to input text and quality of the generated 3D interactions by a very large margin.}
\label{tab:text-to-3d-eval}
\end{table}

% \subsubsection{Qualitative Evaluation}
% \label{subsubsec:3d_qualitative}
\paragraph{Qualitative comparison.}
In \cref{fig:qualitative_compare}, we present representative examples comparing our method with baselines. As shown in rows 1 and 3, Interfusion suffers from Janus artifacts, resulting in multiple hands and missing faces. Even when the generated results appear plausible (row 2), the overall quality is low and the contacts are incorrect.

% In \cref{fig:qualitative_compare}, we show representative examples comparing our method against baselines. It can be seen that SDS-based InterFusion suffers from the Janus problem \TF{I think this is too specific to put in the text here. Better to move it into the description and point out the exact row / col.}, generating blurry 3D results that do not follow the contact descriptions. In several cases, the Janus Problems is severe enough \TF{Is there a severity level to the Janus problem? I thought this is more like a binary thing. Either it has it, or not.} to cause complete collapse of the final generation (\Cref{fig:qualtative_compare} row 1).

TRELLIS, trained natively with 3D data, generates much higher-quality 3D results. However, it is not interaction-aware. The model either generates partial objects (\cref{fig:qualitative_compare}, row 2) or completely omits object generation (\cref{fig:qualitative_compare}, rows 1 and 3).

In contrast, after guiding Sana with our annotated high-quality interaction data, our model is able to precisely follow the text prompt and generate accurate interactions with coherent contacts. Note also how well our model generalizes to different characters, clothes, and hairstyles, while the training data ProciGen contains only 100 human subjects. This clearly shows the advantage of our high-quality data and the great potential of existing models: interaction capability is there, we just need to distill them from structured data.

\subsection{2D Interaction Generation}\label{subsec:2d_baseline}

To obtain high-quality 3D generation with correct contacts, we fine-tune a 2D image generation model on our curated dataset. We evaluate this by comparing our model against the pretrained baseline using the CLIP score, the  GPT score for text consistency, the GPT score for quality, and for contact accuracy. CLIP and GPT scores are computed directly on generated 2D images, while contact accuracy is measured by applying our 3D interaction generation and segmentation pipeline to extract 3D HOI and human registrations from the generated images. As shown in \Cref{tab:text-to-2d-eval}, our method substantially improves contact accuracy, reflecting enhanced interaction awareness in the 2D generation model. Our model also achieves significantly better GPT scores, demonstrating more faithful adherence to input prompts.

In Contact score, our model achieves an accuracy of 90\% in comparison to 45.76\% of the base model. Notably, removing prompts involving \emph{Right Hand} and \emph{Both Feet} causes the baseline score to drop to 23.07\%, whereas our method only decreases slightly to 87.5\%, indicating the base model’s bias toward a limited set of contact configurations. The detailed results are provided in the Supplemental material.

Although our model attains a slightly lower CLIP score than the pretrained model, we attribute this to CLIP’s limited sensitivity to fine-grained physical interactions~\cite{kang2025clipidealnofix, yuksekgonul2023visionlanguagemodelsbehavelike}, as illustrated in \Cref{fig:clip_fail}. 
In contrast, the gains in contact accuracy and text consistency reflect stronger interaction understanding beyond what CLIP can capture.

\begin{table}[t]
\centering
\footnotesize
\setlength{\tabcolsep}{4pt}
\begin{tabular}{l|ccc|c}
\toprule
\multirow{2}{*}{Method} & \multicolumn{3}{c|}{Text Consistency} & \multicolumn{1}{c}{Generation Quality} \\ 
& GPT~$\uparrow$ & CLIP~$\uparrow$ & Contact~$\uparrow$ & GPT~$\uparrow$ \\
\midrule
SANA & 0.31 & \textbf{0.42} & 45.76\% & 0.24 \\
\textbf{Ours} & \textbf{0.69} & 0.40 & \textbf{90\%} & \textbf{0.76} \\
\bottomrule
\end{tabular}
\caption{\textbf{Quantitative comparison with 2D baselines.} We compare our method against pretrained text-to-2D generation SANA~\cite{xie2024sana} model. Our method achieves higher text consistency and 3D interaction quality. The CLIP score is however not very informative due to its limited sensitivity to fine-grained text \cite{kang2025clipidealnofix, yuksekgonul2023visionlanguagemodelsbehavelike} as we also show in~\cref{fig:clip_fail}.}
\label{tab:text-to-2d-eval}
\end{table}

\subsection{Ablation Studies}
\label{subsec:ablation}
We ablate different design choices of our method. 
% \subsubsection{Pipeline Components}
\paragraph{High-Quality Interaction Data.}
A critical step in our data curation pipeline is filtering undesirable data with issues such as interpenetration, implausible actions, and mismatched action-contact pairs. We compare our model against one trained on the entire ProciGen dataset in \cref{tab:Ablations}b. While the full ProciGen dataset provides diverse interaction examples, the aforementioned issues cause the model to learn incorrect and implausible actions. Consequently, our model trained on filtered data points achieves higher contact accuracy and GPT scores.

% to show that filtering improves both contact quality and visual fidelity. We also ablate our retexturing choice to demonstrate that, while retexturing enhances visual quality, it does not affect contact scores.

% For this section, we use the same settings as in \Cref{subsec:3d_baseline}, but when calculating GPT-Score, we provide ChatGPT with three generations from three views of each model.

% As shown in \Cref{tab:Ablations}, our model trained on only 400 examples outperforms the model trained on the full ProciGen dataset in both contact scores and visual quality. 

\paragraph{Retexturing.} We also observe that removing retexturing causes a small drop in contact scores, but significantly reduces GPT-Score, as can be seen in \cref{tab:Ablations}c. This indicates that retexturing is essential for better texture quality.

\begin{table}[h!]
\centering
\footnotesize
\setlength{\tabcolsep}{3.5pt}
\begin{tabular}{lccc}
\toprule
\multirow{2}{*}{Method} & \multicolumn{3}{c}{Text Consistency} \\
& GPT~$\uparrow$ & CLIP~$\uparrow$ & Contact accuracy~$\uparrow$ \\
\midrule
\textbf{a. Ours w/o data filter} & 0.20 & 0.413 & 0.80 \\
\textbf{b. Ours w/o retexturing} & 0.05 & 0.412 & 0.85 \\
\textbf{c. Ours} & \textbf{0.75} & \textbf{0.417} & \textbf{0.90} \\
\bottomrule
\end{tabular}
\caption{\textbf{Ablation studies.} Our proposed data filtering improves contact accuracy and retexturing improves consistency to text input (GPT score). Combining both achieves the best result.}
\label{tab:Ablations}
\end{table}

\paragraph{View conditioned sampling.}
To produce the final 3D model with correct contacts, we rely on a pretrained image-to-3D model. 
% We exploit the interaction priors from 3D foundation models by providing it 2D images with improved interaction quality and consistency. We rely on a pretrained image-to-3D generation model~\cite{hunyuan3d22025tencent} to obtain 3D interaction instead of fine-tuning it due to the lack of high-quality 3D HOI data. 
This image to 3D lifting is mostly accurate when both the human and object are visible. To make the 3D generation more stable, we propose to condition the 2D image sampling on a text that specifies which view to generate. To evaluate this, we compute the contact accuracy of generating one, two or three views, respectively. When more than one views are generated, the contact accuracy is simply the maximum of the generated views. We send the ground-truth image renderings from ProciGen and also the images generated by our view conditioned model to Hunyuan3D and report the contact accuracies in \cref{tab:view-cond-sample}. It can be seen that Hunyuan3D can preserve the contacts over 79\% of the time when only one view is given. However, it faithfully preserves the contacts with more than 94\% accuracy when we sample three times with different views. This highlights the importance of our view conditioned sampling.

\begin{table}[h!]
\centering
\begin{tabular}{l|ccc}

Method &3 views & 2 view & 1 view \\
\hline
ProciGen~\cite{xie2023template_free} images &  94.7\% & 89.5\% & 79.6\%  \\
Our generated images & 90.0\% & 86.7\% & 78.3\%\\ 
\hline
\end{tabular}
\caption{\textbf{Contact accuracy} of the lifted 3D given different number of 2D images. Our view-conditioned sampling allows generating three views conditioned on the view description, leading to more stable 3D results.}
\label{tab:view-cond-sample}
\end{table}

\paragraph{Human registration.} We propose a simple yet effective pipeline to register SMPL model to the generated interaction mesh. We compare this against ETCH~\cite{li2025etch}, a state-of-the-art human registration method. We compute the one-directional Chamfer distance from the segmented human mesh to the registered SMPL mesh, and the results are: 4.30cm (ETCH) vs. 1.60cm (ours). ETCH was trained on scans with mainly standing poses, hence it cannot handle complex poses like sitting and bending. Our method is training-free and overall more robust to different poses. 

% \begin{table}[]
%     \centering
%     \begin{tabular}{c|c}
%     Method & Chamfer distance \\
%     \hline
%          ETCH \cite{li2025etch}& 0.043 \\
%         Ours & \textbf{0.016}
%     \end{tabular}
%     \caption{Evaluating our proposed human registration. We are better than ETCH.}
%     \label{tab:human-reg}
% \end{table}

\section{Limitation and Future Work}
Despite that our method can follow detailed text prompts and generate coherent 3D interactions, one limitation that we noticed is that while our model can reason about the contacts very well,  it has difficulties with following very complex human pose descriptions. Notably, such human poses can be very ambiguous when described in text. As opposed to body part contacts, it is also difficult to obtain unique examples demonstrating characteristic poses. Therefore, future work could introduce a dedicated model for text-to-human-pose generation to further improve the pose awareness, similar to InterFusion\cite{dai2024interfusion} or ChatPose\cite{feng2024chatpose}.

\section{Conclusion}
In this paper, we address the challenging problem of generating 3D-human-object-interactions from detailed text prompts by leveraging existing foundation models for curating diverse and high-quality training data, and to establish an accurate text-to-3D pipeline. 
We show that little data for interaction is enough to adjust the rich representations learned in general text-to-image models, and that with view-conditioning, the text-to-image output can  be lifted to high-quality 3D meshes with accurate contact. Our results outperform the baselines by nearly an order of magnitude, exhibiting strong generalization to diverse categories and interaction types.

\noindent
{\footnotesize
\textbf{Acknowledgements.} We thank RVH members \cite{rvh_grp} for their helpful discussions. This work is funded by the Deutsche Forschungsgemeinschaft (German Research Foundation) - 409792180 (Emmy Noether Programme,
project: Real Virtual Humans), and German Federal Ministry of Education and Research (BMBF): Tübingen AI Center, FKZ: 01IS18039A, and Amazon-MPI science hub. Gerard Pons-Moll is a Professor at the University of Tübingen endowed by the Carl Zeiss Foundation, at the Department of Computer Science and a member of the Machine Learning Cluster of Excellence, EXC number 2064/1 – Project number 390727645.
}

{
    \small
    \bibliographystyle{ieeenat_fullname}
    \bibliography{main}
}
% \clearpage
\clearpage
\setcounter{page}{1}
\maketitlesupplementary

In this supplementary, we provide further details about our method implementation in \cref{sec:implementation} and experiment setup in \cref{sec:exp-setup}. We then show additional results to demonstrate the controllability and generalization of our method in \cref{sec:more-results}. Our code and models will be fully released to enable easy reproduction of our results. 

\section{Implementation Details}
\label{sec:implementation}

\subsection{Segmentation and Reconstruction}

In this Section we provide more details of the semantic separation of the human-object-interaction mesh.
To separate the single watertight mesh, employ a three-stage mesh separation pipeline that relies on 1) controlled multi-view rendering, 2) open-vocabulary video segmentation, and 3) a per-vertex voting strategy from geometric consistency checks.

\textbf{Multi-View Rendering.} To obtain high-quality segmentation in the later stage, the rendering has to be of \textit{high quality}, \textit{clearly show the object}, and \textit{have a smooth trajectory}.
For high quality renders of our mesh, we use the open-source rendering engine of TRELLIS~\cite{xiang2024trellis} and render $120$ views per object.
To facilitate high quality segmentation, we construct the camera trajectory as a multi-band spherical sweep around the mesh, which provides a broad view coverage and smooth viewpoint transitions.
We partition the full set of views into several elevation bands from $[-60^\circ,60^\circ]$ and perform a full $360^\circ$ azimuthal sweep at a fixed elevation.
To make the transition to the next elevation smooth, the azimuth direction alternates after each band.

All views are rendered at a constant distance and the same field of view, ensuring that the object stays at a consistent scale.
After generating the full set of views, we cyclically rotate the sequence so that it begins at a diagonally elevated viewpoint, which we found to already give good enough segmentation quality.
Otherwise, one could also run detections on each frame and chose the starting point based on the maximum confidence, however, we found this not to be neccessary.
For each view, we then store the RGB image, the rendered depth map, and the camera transformation.

\textbf{Open-Vocabulary Video Segmentation.} 
The rendered RGB sequence is processed using Grounded Segment Anything 2 (GSAM2). We query the model twice: once using the text prompt “person” and once using the known object category. This yields two temporally coherent binary mask sequences, one for the human and one for object.

\textbf{Vertex-Level Labeling and Mesh Separation.}
Given access to the depth map $\mathbf{D}_i$ and camera parameters for each view $i$, we project each mesh vertex $\textbf{v}$ into all frames and record the views for which the vertex is 1) inside the image bounds and 2) passes a z-buffer consistency check to determine the closest vertex along the camera ray.

For each vertex, we inspect the object mask values at its projected location across all visible views. 
We compute the fraction of visible views where the vertex is marked as object.
A vertex is then assigned to the object class, if this fraction exceeds a pre-defined threshold $\tau$ (in our case, we simply chose $\tau=0.5$).
Each triangle is considered part of the object, if two out of three connected vertices have been labeled as object.
Finally, vertices and triangles are partitioned accordingly to obtained labels to obtain object mesh $\mathcal{O}$ and the remaining vertices and triangles form the human mesh $\mathcal{H}_m$.

% \textbf{Optional Watertight Surface Reconstruction.}
% The vertex-wise separation of the combined mesh naturally introduces gaps along the human–object interface. 
% For downstream tasks like animation, it could be desirable to have watertight meshes to work with.
% Since the holes in the meshes are highly localized, the missing surface can be reconstructed reliably using Poisson surface reconstruction~\cite{kazhdan2006poisson}.
% We showcase some examples where we used PyMeshlab~\cite{meshlab} with the default parameters in ~\Cref{fig: reconstruct}, however for higher quality meshes one could use a deep learning based surface reconstruction method.

% \begin{figure*}[t]
%     \centering
%     \begin{overpic}[width=0.9\linewidth]{Figures/segmentations_suppl.pdf}
%         \put(8,-1){\ Full Mesh}
%         \put(22,-1){ Human Mesh $\mathcal{H}_m$}
%         \put(41,-1){ Object Mesh $\mathcal{O}$}
%         \put(60,-1){Reconstruction}  
%     \end{overpic}
    
%     \caption{Qualitative samples of our semantic mesh separation pipeline.  Each column shows the full mesh, human mesh $\mathcal{H}_m$, object mesh $\mathcal{O}$, and the surface reconstruction using meshlab~\cite{meshlab, kazhdan2006poisson}, respectively.
%     }
%     \label{fig: reconstruct}
% \end{figure*}

\subsection{SMPL fitting for base model}
A common issue with the base image generation model is that it often generates only partial humans: sometimes only the torso is visible, sometimes only the lower body, and sometimes certain limbs are missing. Since CameraHMR always returns a full SMPL mesh $\mathcal{H}_s$, the Chamfer alignment fails due to mismatched scale and wrong matches.

To address this, we obtain a partial SMPL mesh $\mathcal{H}'_s$ that corresponds to the visible human regions. To obtain this partial mesh, we first use Grounded SAM 2 to compute a human mask $\mathbf{M}^{h}_{front}$ for the front-facing view of the 3D model (recall that CameraHMR can be utilized to calculate the front view). We then subset the SMPL vertices that fall inside this mask.

Formally, for an SMPL mesh with vertex set $\mathcal{V}$, we define the subset $\mathcal{V}'$ as:
$$\mathcal{V}' = \{\, \mathbf{v} \in \mathcal{V} \mid \mathbf{M}_{\text{front}}^{h}\!\left[\pi_{\text{front}}(\mathbf{v})\right] = 1 \,\}$$
where $\pi_{\text{front}}$ denotes the camera projection for the front view. The vertices in $\mathcal{V}'$ define a partial SMPL mesh that can be reliably aligned with the partial generated mesh.

Finally, once the alignment is computed, we apply the transformation from the CameraHMR coordinate system to the mesh coordinate system, $\mathbf{T}_{\text{camHMR} \rightarrow \text{mesh}}$, to the full SMPL mesh so that accurate contact computation can be performed.

\subsection{Animation}
Given the SMPL mesh $\mathcal{H}_s$ with pose and shape parameters $\theta$ and $\beta$, respectively, we copy only the pose parameters from the animation sequence while keeping the original shape fixed.

Because the animation is performed in the unaligned SMPL coordinate space, we first transform both the human mesh $\mathcal{H}_m$ and the object mesh $\mathcal{O}$ using the transformation $\mathbf{T}_{\text{mesh} \rightarrow \text{camHMR}}$, which is the inverse of the alignment transform that maps the SMPL mesh to the segmented human mesh.

Next, we transfer the linear blend skinning (LBS) weights by finding, for each vertex in transformed $\mathcal{H}_m$, its nearest neighbor in $\mathcal{H}_s$. The LBS weights from the corresponding vertex in $\mathcal{H}_s$ are then assigned to the vertex in $\mathcal{H}_m$.

\begin{figure*}[ht]
    \centering
    \includegraphics[width=1.0\linewidth]{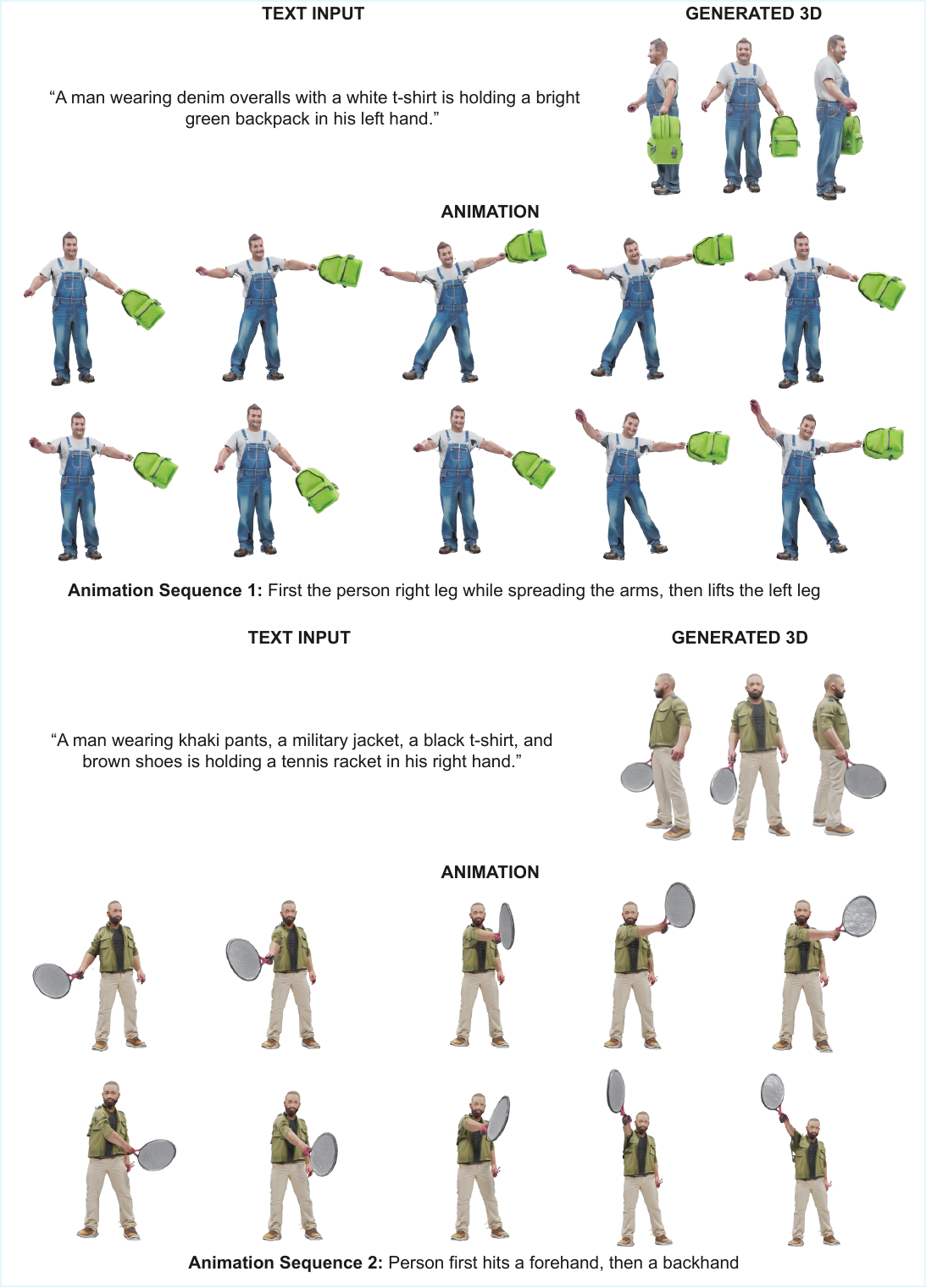}
    \caption{
    \textbf{Animation results.} Our fitted SMPL and segmented objects allow reanimation of the generated human object interaction mesh
    }
    \label{fig:animation}
\end{figure*}

For the object mesh, we attach it to the nearest SMPL joint. If multiple joints have similar distances to the object, we randomly choose one of them as the attachment point. Examples of animation can be found at \Cref{fig:animation} 

\section{Experiment Details}\label{sec:exp-setup}
\subsection{More Details On Experimental Setup}
For image-to-shape generation, we use \textit{Hunyuan3D-DiT-v2.0} for geometry and \textit{Hunyuan3D-Paint-v2.1} for texture synthesis. The \textit{2.0} model provides more stable geometric results in our experiments, while \textit{2.1} yields higher-quality, PBR-compatible textures. 

For Grounded-SAM, we use \textit{sam2.1-hiera-large.pt} for the detection component and \textit{grounding-dino-tiny} to generate the initial candidates. The confidence threshold for GroundingDINO is empirically set to 0.4.

\subsection{View Conditioned Sampling}
\begin{figure}[ht]
    \centering
    \includegraphics[width=1.0\linewidth]{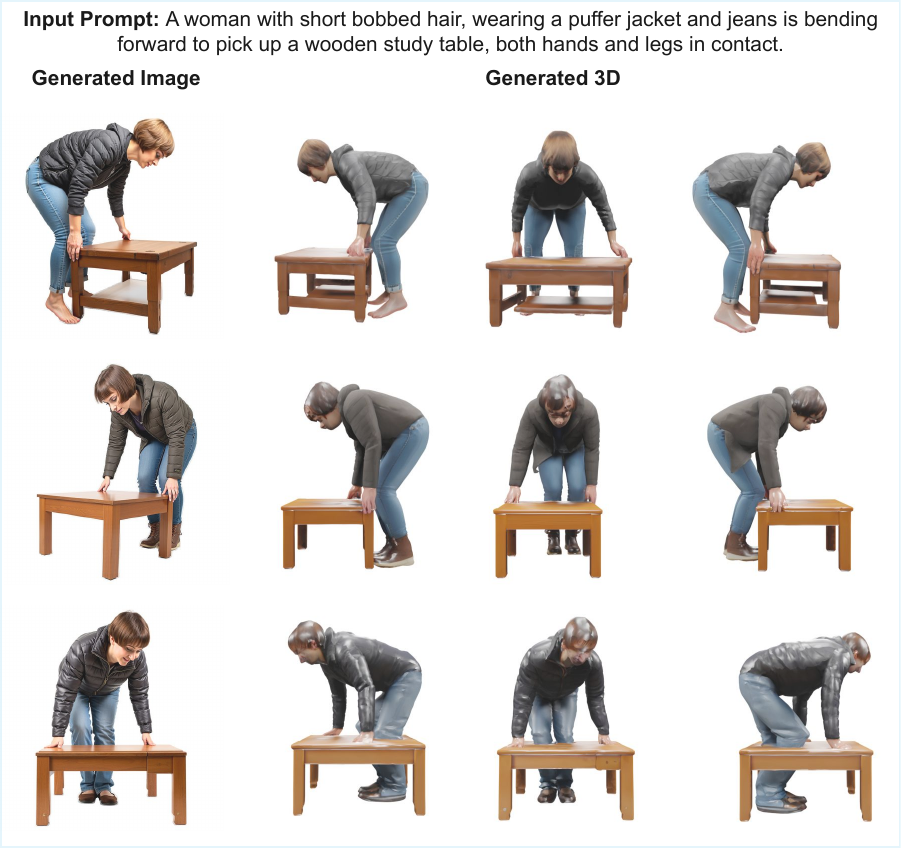}
    \caption{
    \textbf{Advantage of view conditioned sampling.} Given same interaction prompt, our method generates three views that all correctly follow the contacts. Yet Hunyuan3D stuggle to reason the compositional shape under occlusion such as the leg occluded by the table in front view. By sampling side views as input to Hunyuan3D, we are able to generate at least one plausible 3D human-object interaction for each text prompt. 
    % Even when all 3 generated images are correct, the front view image causes occlusion which leads to incorrect generation where human severely penetrates the object.
    }
    \label{fig:view_qual}
\end{figure}

In Section 4.4 of the main paper, we qualitatively demonstrated that view conditioning is essential for producing correct contacts. In \Cref{fig:view_qual}, we further illustrate this effect. Although all three 2D images are correct, in the front view image (third row), occlusions lead to failures during 3D lifting. This highlights that multiple views are critical for obtaining accurate 3D generations.

\subsection{GPT Score}
\begin{figure}[ht]
    \centering
    \includegraphics[width=1.0\linewidth]{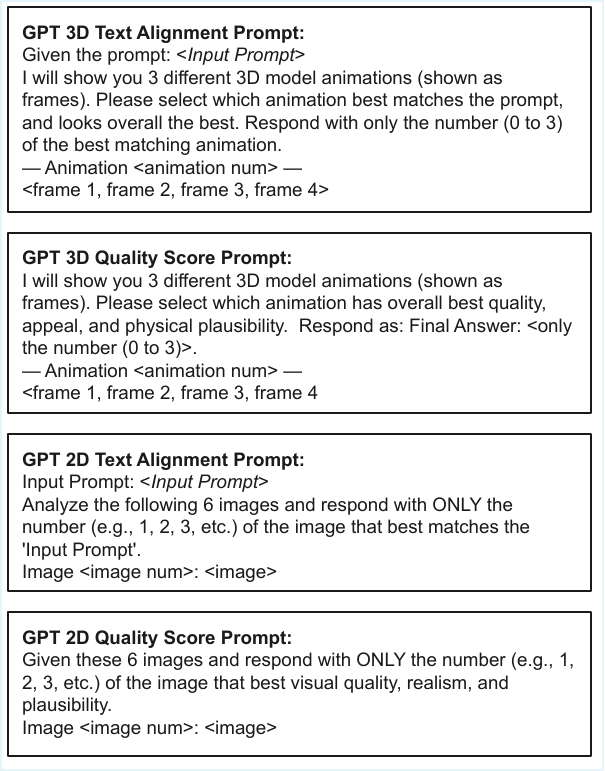}
    \caption{
    \textbf{Prompts used for calculating GPT scores.} We instruct GPT-4V to evaluate the text alignment and quality for generated 3D models (rendered as video) or 2D images. 
    }
    \label{fig:gpt_score}
\end{figure}

We provide here the prompts used to compute the GPT Score shown in \Cref{fig:gpt_score}. For the 3D scores, we randomly select one of three available views for our case. For the 2D scores, we sample both the base model and our model three times to ensure a fair comparison.

One important detail is that GPT has limited understanding of fine-grained contact information. As a result, explicit contact points in the user prompts can introduce ambiguity. To mitigate this, we replace specific contact descriptions with more generic phrasing. For example, a prompt such as \emph{“holding a bag in the left hand”} is changed to \emph{“holding a bag in one hand”.} We hence evaluate the consistency is a less strict way. 

\subsection{User Study}
\begin{figure*}[ht]
    \centering
    \includegraphics[width=1.0\linewidth]{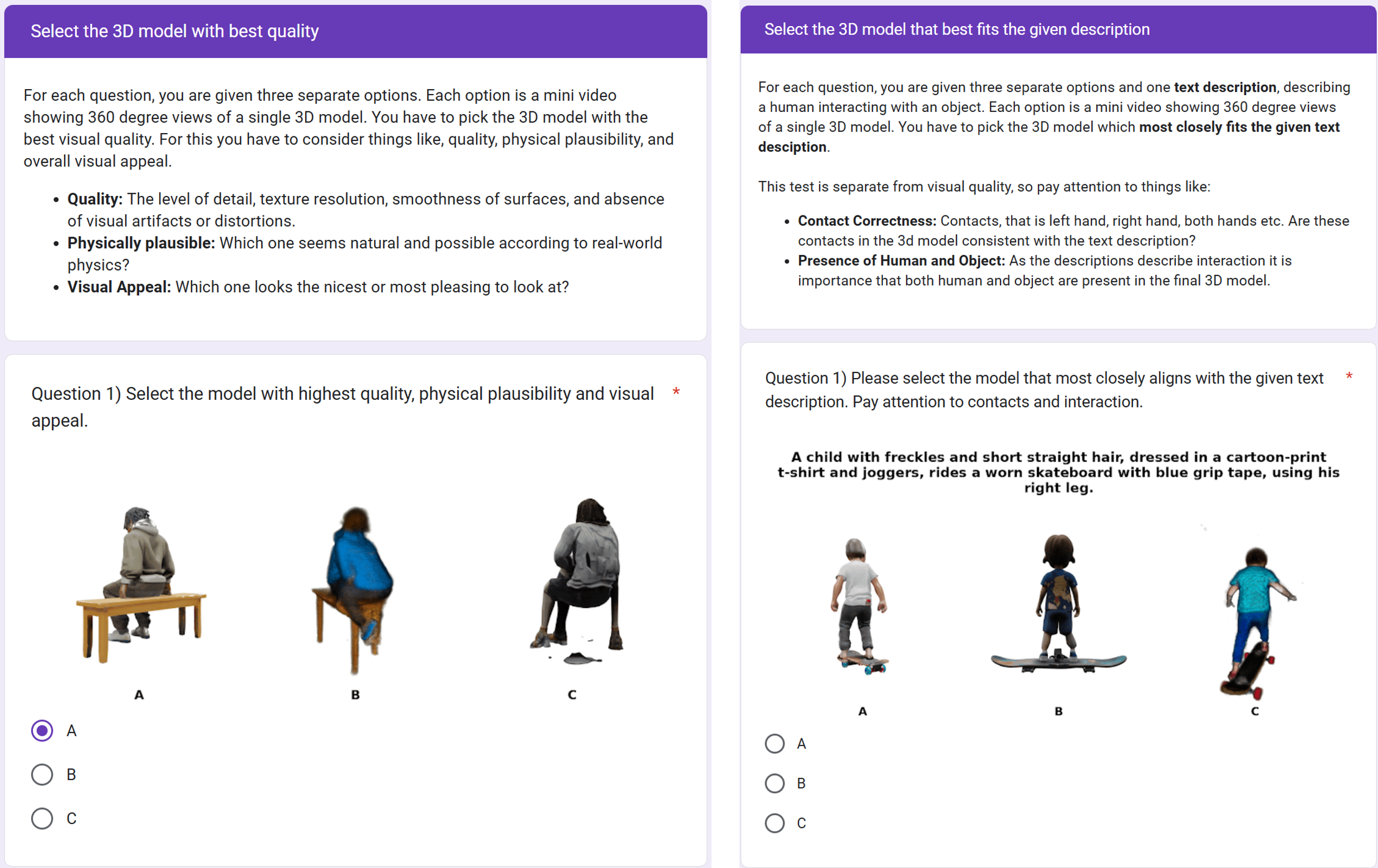}
    \caption{
    \textbf{User study instruction and example questions.} We guide participants to evaluate the quality or consistency to input text prompt. We randomly select 20 examples for each aspect and users are asked to choose the best one from three options. 
    }
    \label{fig:user_study}
\end{figure*}

In \Cref{fig:user_study}, we show the instruction screenshots used for our user study. We ask participants to assess \emph{quality} first so that they are not biased to select human object interaction images, and actually focus on the quality related details. Then, in the second part, we explicitly specify our task that is human object interaction generation.

\subsection{Contact Accuracy}

In \Cref{tab:part_contact_score}, we demonstrate that our model generalizes well across diverse contact configurations, whereas the base model exhibits a strong bias toward only a few of them. Notably, for the \emph{Left Leg} configuration, the base model produced extremely poor generations for one of the prompts, leading to a complete failure of the 3D lifting stage. To maintain a fair comparison, we report the average score over the remaining nine successful generations. Even under this favorable evaluation, the base model achieves correct contact only 11.11\% of the time.

\begin{table}[t]
\centering
\footnotesize
\setlength{\tabcolsep}{3pt}
\begin{tabular}{lcc}
\toprule
\textbf{Body Part} & \textbf{SANA} & \textbf{Ours} \\
\midrule
Left Hand ($\uparrow$)  & 40\%   & \textbf{100\%} \\
Right Hand ($\uparrow$) & 90\%   & \textbf{100\%} \\
Both Hands ($\uparrow$) & 10\%   & \textbf{100\%}  \\
Left Leg ($\uparrow$)   & 11.11\% & \textbf{80\%} \\
Right Leg ($\uparrow$)  & 30\%   & \textbf{70\%} \\
Both Legs ($\uparrow$)  & 90\% & \textbf{90\%} \\
\bottomrule
\end{tabular}
\caption{\textbf{Per-part contact accuracy.} 
Our model generates correct contacts for various contact scenarios whereas base model SANA~\cite{xie2024sana} can follow mainly `right hand' and `both legs' but fails in other body parts.}
\label{tab:part_contact_score}
\end{table}

\section{Additional Results}\label{sec:more-results}
\subsection{Controllability}
\begin{figure*}[ht]
    \centering
    \includegraphics[width=\textwidth]{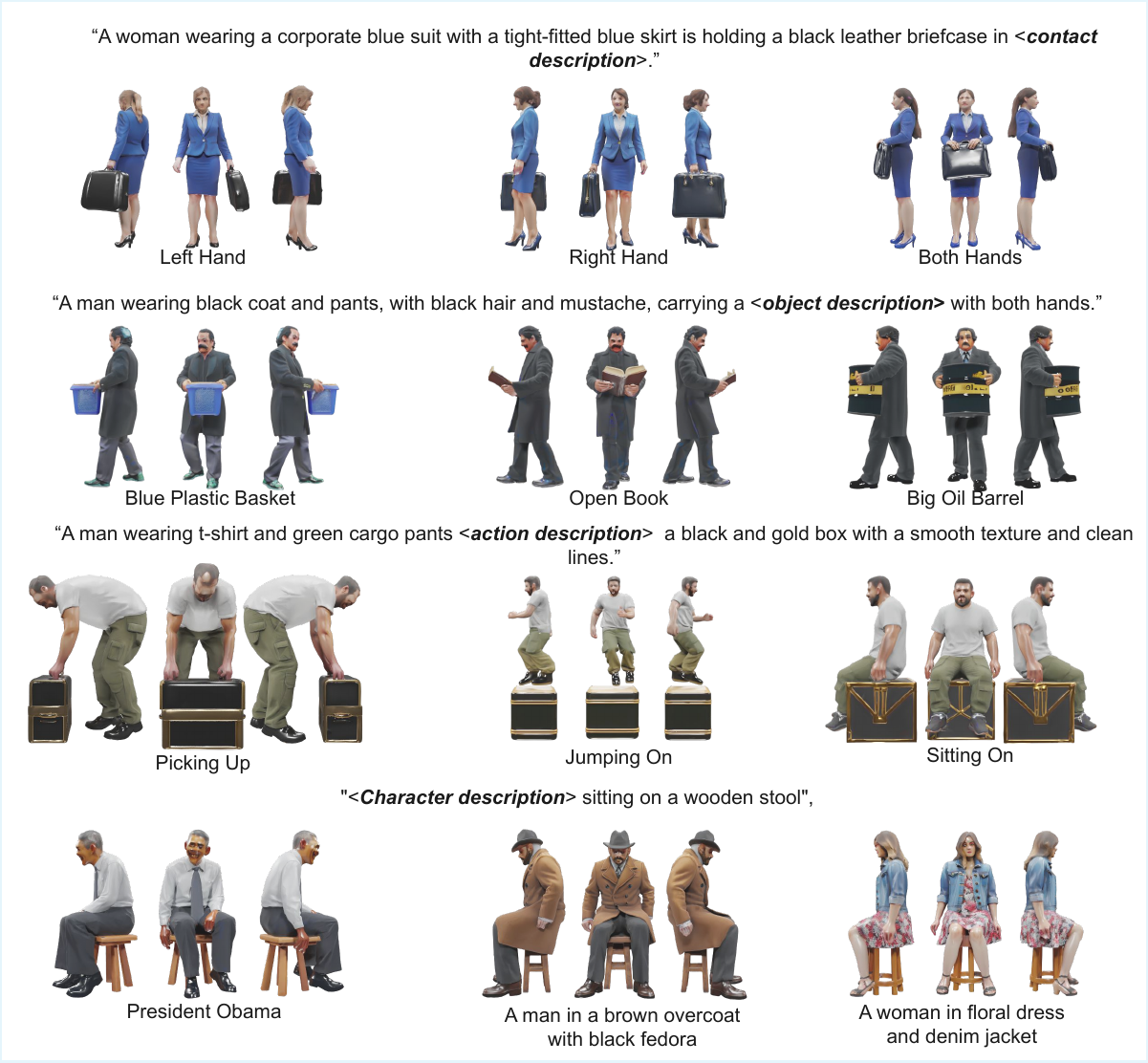}
    \caption{\textbf{Controllable interaction generation via text.} We can control the contact, object, action or human description in our text prompt with minimum changes to the other parts. This makes it possible to use our model as a data generator.}
    \label{fig:controllability}
\end{figure*}

Through our fine-grained text annotation and data filtering, we successfully enhance the interaction awareness of image generation model Sana. Interestingly, it also learns the decoupling of key components for interaction. We show in \cref{fig:controllability} that one can change the text description of human, object, action label, or contact regions. After which the model precisely follows the new text description while keeping the other parts barely untouched. This shows the superior text following capability of our model and makes it possible to repurpose our model as an interaction data generator. 

\subsection{OOD Generalization}
\begin{figure*}[ht]
    \centering
    \includegraphics[width=1.0\linewidth]{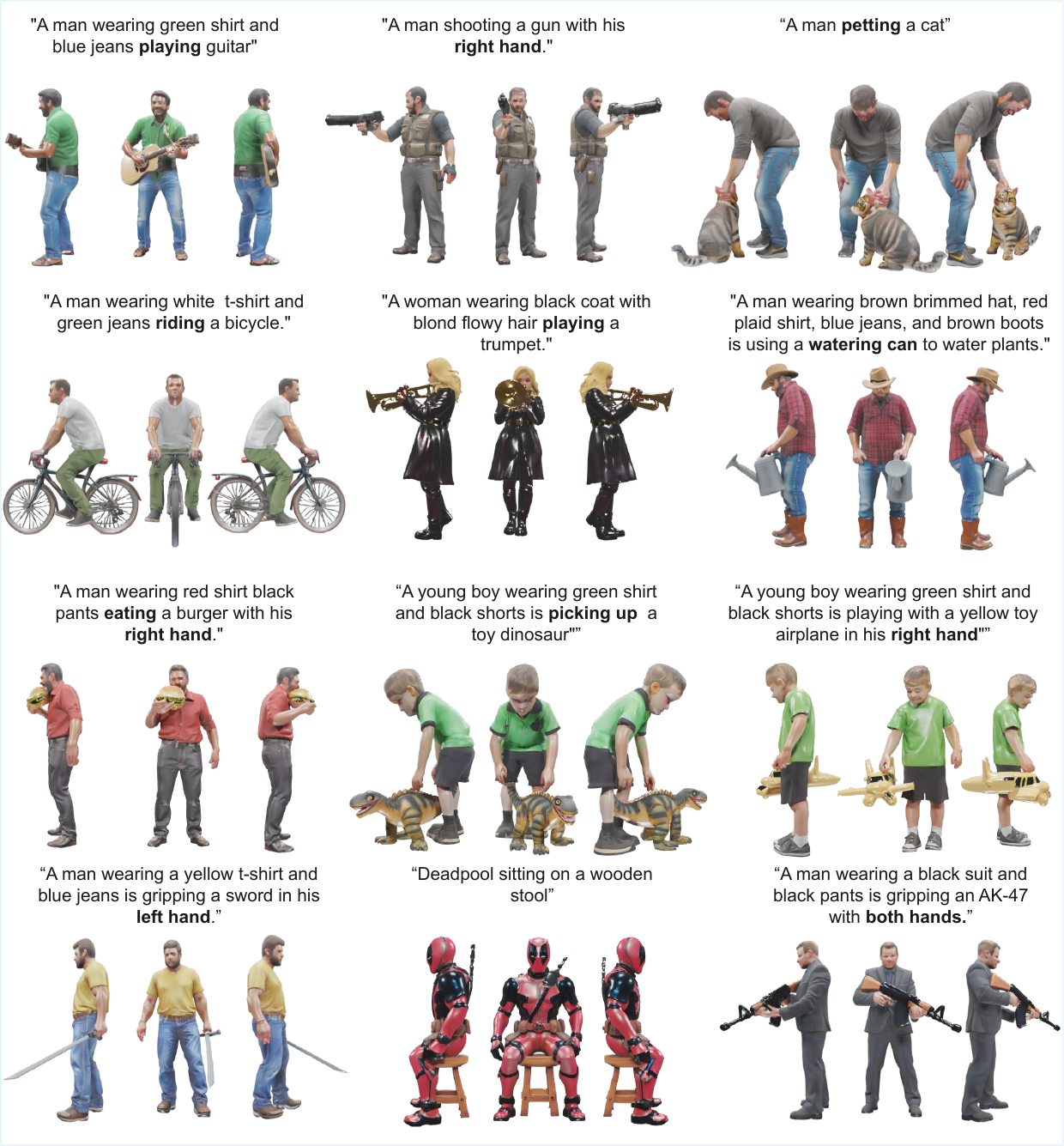}
    \caption{
    \textbf{OOD generalization.} Fine tuned only on interaction data with 100 humans and 15 object categories, our method generalizes well to different objects, human descriptions and new actions.
    }
    \label{fig:ood_generalization}
\end{figure*}

Although our fine-tuning set contains only ~400 images and we train for roughly 1050 epochs, raising potential concerns about overfitting, our results show otherwise. As illustrated in \Cref{fig:ood_generalization}, our model not only generalizes robustly to previously unseen subjects, but also synthesizes plausible interactions with out-of-distribution objects and can generate coherent out-of-distribution actions.

\subsection{More Qualitative comparison}

\begin{figure*}[ht]
    \centering
    \includegraphics[width=1.0\linewidth]{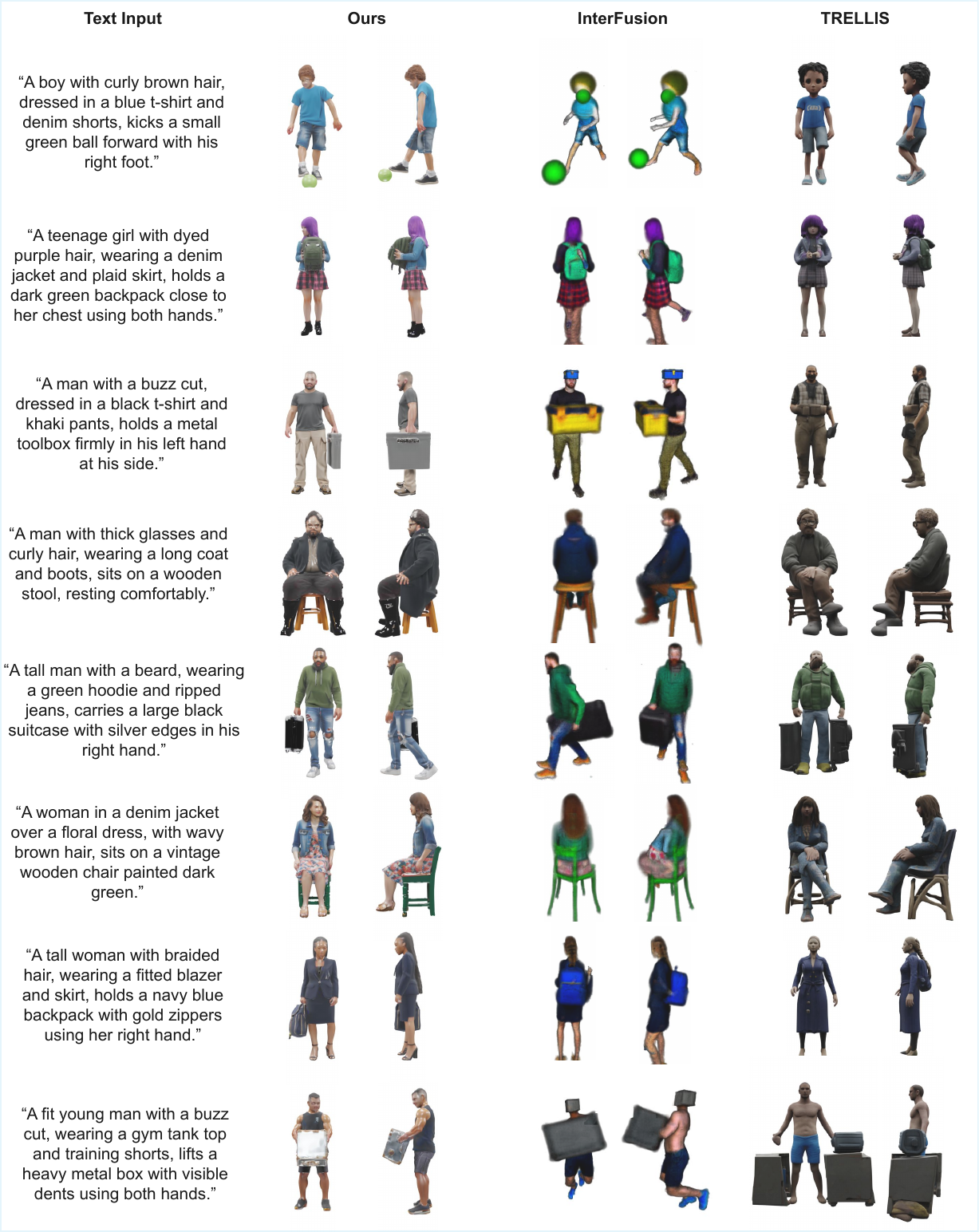}
    \caption{
    \textbf{More qualitative comparison.} Our method consistently produces high quality results with correct contact and details.
    }
    \label{fig:more_qual}
\end{figure*}

In \Cref{fig:more_qual}, we provide additional qualitative comparisons with current 3D baselines. A key advantage of our method is its reliability at inference. InterFusion, which relies on score distillation sampling, requires users to manually fine-tune parameters for each new generation. In contrast, our approach only needs to be fine-tuned once after which the same network generalizes to different 3D outputs.

\subsection{Text Annotation Examples}

\begin{figure*}[ht]
    \centering
    \includegraphics[width=1.0\linewidth]{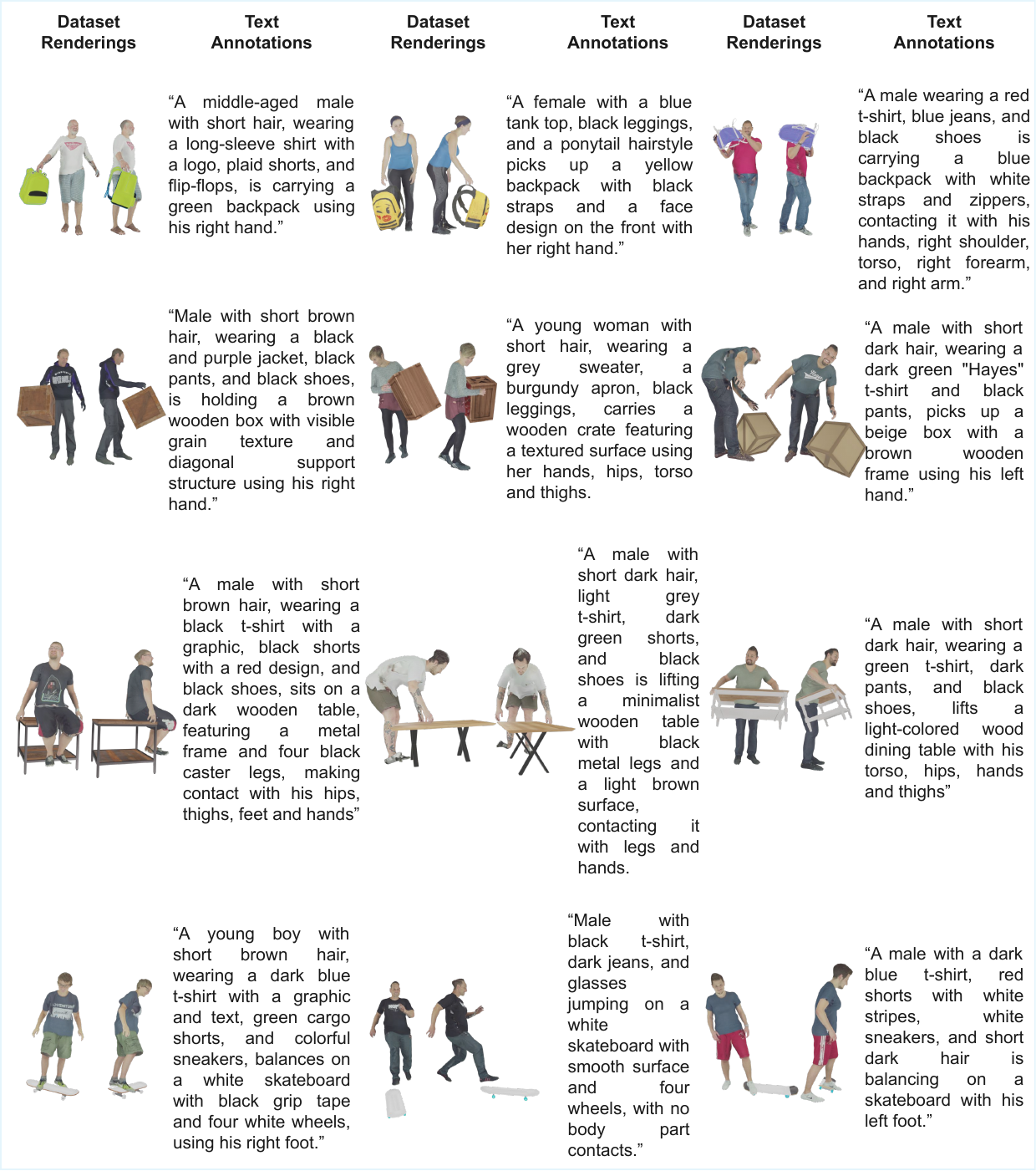}
    \caption{
    \textbf{More examples from our annotated data.} Our annotation pipeline generates detailed descriptions about human, object, interaction action and contacts.
    }
    \label{fig:example_anno}
\end{figure*}

In the \Cref{fig:example_anno}, we show a few examples from our annotation pipeline. Our decomposed annotations, allows accurate annotation for human, object, contact and interaction. These annotations, help us in subsequent filtering and effective fine-tuning.

\end{document}